\documentclass[
]{ceurart}

\usepackage{listings}
\usepackage{mathptmx}
\usepackage{multirow}
\usepackage{comment}
\usepackage{hyperref}
\usepackage{tablefootnote}
\usepackage{xcolor}
\usepackage{subcaption}
\usepackage{amsmath}

\usepackage{graphicx}
\usepackage{amsmath}
\usepackage{booktabs}
\begin{document}

\copyrightyear{2024}
\copyrightclause{Copyright for this paper by its authors.
  Use permitted under Creative Commons License Attribution 4.0
  International (CC BY 4.0).}

\conference{HHAI-WS 2024: Workshops at the Third International Conference on Hybrid Human-Artificial Intelligence (HHAI),
June 10—14, 2024, Malm\"o, Sweden}

\title{Multi-Perspective Stance Detection}

\author[1,2]{Benedetta Muscato}[
email=benedetta.muscato@sns.it,
orcid=0009-0000-3061-8408
]
\cormark[1]
\address[1]{Scuola Normale Superiore, Italy}
\address[2]{University of Pisa, Italy}

\author[3,2]{Praveen Bushipaka}[
email=praveen.bushipaka@phd.unipi.it,
orcid=0009-0009-7753-8662
]
\address[3]{Scuola Superiore Sant'Anna,Italy}

\author[1]{Gizem Gezici}[
email=gizem.gezici@sns.it,
orcid=0000-0001-9782-5751
]

\author[2]{Lucia Passaro}[
email=lucia.passaro@unipi.it,
orcid=0000-0003-4934-5344
]

\author[1]{Fosca Giannotti}[
email=fosca.giannotti@sns.it,
orcid=0000-0003-3099-3835
]


\cortext[1]{Corresponding author.}

\begin{abstract}
  Subjective NLP tasks usually rely on human annotations provided by multiple annotators, whose judgments may vary due to their diverse backgrounds and life experiences. Traditional methods often aggregate multiple annotations into a single ground truth, disregarding the diversity in perspectives that arises from annotator disagreement. In this preliminary study, we examine the effect of including multiple annotations on model accuracy in classification. Our methodology investigates the performance of perspective- aware classification models in stance detection task and further inspects if annotator disagreement affects the model confidence. The results show that multi-perspective approach yields better classification performance outperforming the baseline which uses the single label. This entails that designing more inclusive perspective-aware AI models is not only an essential first step in implementing responsible and ethical AI, but it can also achieve superior results than using the traditional approaches.
\end{abstract}


\maketitle

\section{Introduction}

Large language models (LLMs) have revolutionized natural language processing (NLP) field by outperforming state-of-the-art approaches in NLP tasks.
%
In the context of supervised learning settings, which require labeled data, it is well-known that human annotators may provide diverse perspectives via annotations due to the intricacies of thought, varied life experiences, and diverse educational backgrounds~\cite{romberg2022your, soni2024comparing}.
This inevitably leads to disagreement especially in the subjective tasks such as toxicity, argumentation mining, stance and hate speech detection in which multiple perspectives may be equally valid, and a unique ‘ground truth’ label may not exist.
To validate this assumption, we intend to revisit the task of stance detection through involving multiple annotations, i.e. diverse perspectives of multiple annotators, instead of a single ground truth. 
In this preliminary study, our contributions are two-fold: (i) we investigate if establishing a more inclusive ethically-aware LLM can provide better classification performance, and (ii) further explore if the model confidence is affected by annotator disagreement.
The main concern related to the traditional aggregation procedures, which reduce multiple annotations into a single label using standard approaches of disagreement resolution through majority voting, is oversimplifying the real-world complications by assuming the presence of a single ground truth~\cite{basile2021we, kanclerz2022if, mokhberian2023capturing}.
Due to these concerns, a novel approach called~\emph{perspectivism} has recently been proposed for gathering annotated data in NLP~\cite{Rottger2021Two}. This new paradigm promotes the use of disaggregated datasets, containing different human judgments, especially for subjective tasks.
This approach holds the potential of fostering the development of~\emph{responsible} and~\emph{ethical} AI systems in which fairness is promoted by incorporating the perspectives of diverse annotators. These perspectives arise not only from different backgrounds but also from different ideas, ensuring a more comprehensive and equitable representation.

\begin{figure}[htbp]
    \centering
    \includegraphics[width=0.9\linewidth]{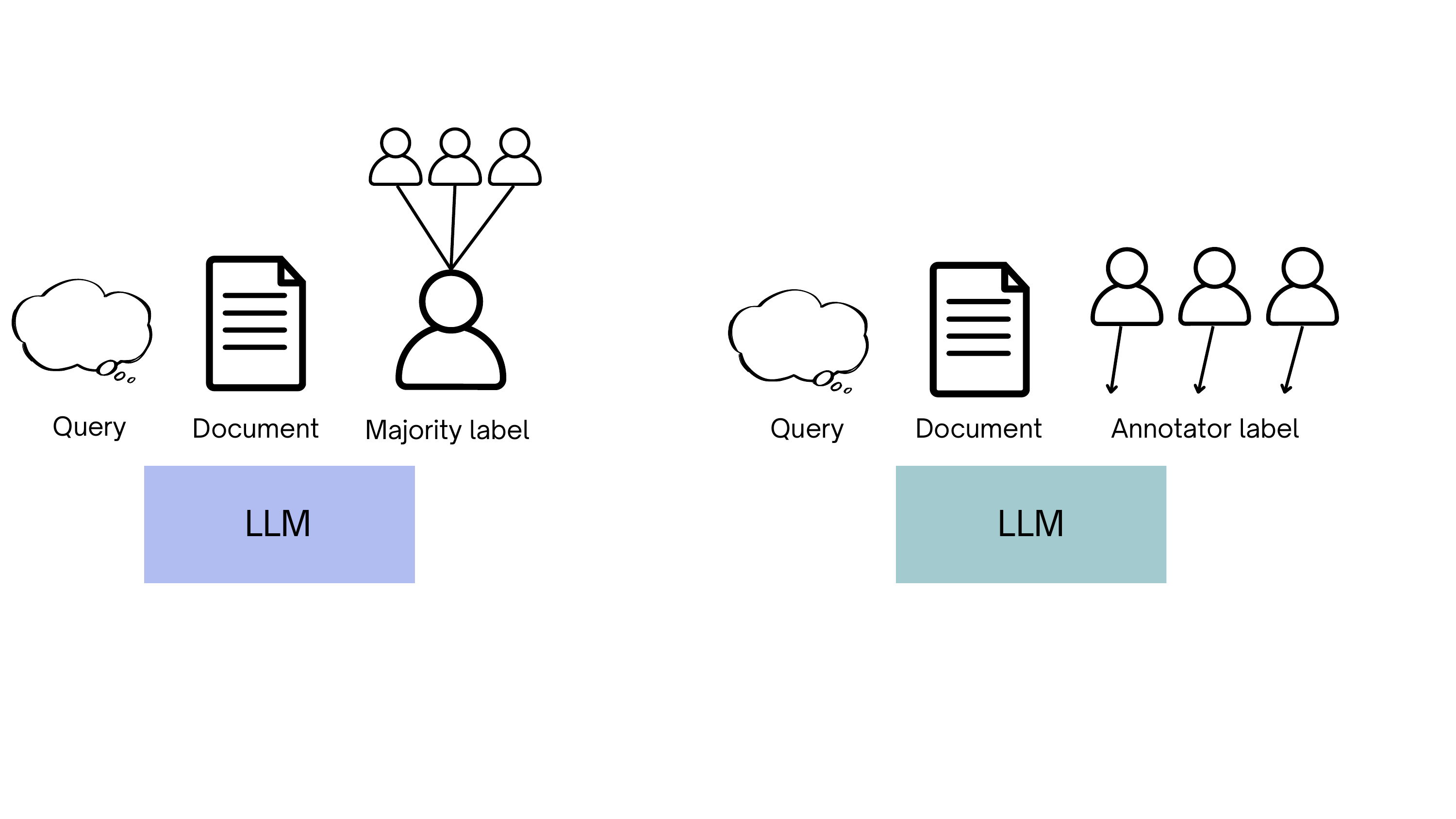}
    \vspace{-3.5em} 
    \caption{Baseline vs. Multi-perspective Approach in Model Finetuning. Baseline relies on aggregated label via majority voting (i.e. Majority label), while Multi-perspective uses each annotator's individual label (i.e. Annotator label).} 
    \label{schema}
    \vspace{-0.5em}  
\end{figure} 
\vspace{-1.5em}

\section{Methodology}
In the following section we describe our methodology to employ multi-perspective stance classification. 

\subsection{Dataset}
We use a previously collected dataset to apply the multi-perspectivist approach to the stance classification task \cite{gezici2021evaluation}.
The dataset is composed of top-10 news documents that have been crawled from the two popular search engines by issuing 57 queries in total which correspond to a wide range of controversial topics; including but not limited to education, health, entertainment, religion and politics. 

After the crawling, the authors further obtained the stance label of each document with respect to the queries via crowd-sourcing using the MTurk Platform \footnote{Amazon Mechanical Turk}. In line with~\citet{gezici2021evaluation}, each document was judged by three crowd-workers. A stance label can have the following values:~\emph{pro, neutral, against, not-about,} and~\emph{link not-working}. Note that due to the nature of the MTurk platform, although there are only three annotations for each document, this does not guarantee that each document has been annotated by the same three annotators. Thus, there is a high probability that the annotated dataset contains more than three diverse perspectives which highly affects the disagreement level.

Both the Fleiss-Kappa score of $0.3500$  and the Inter-Rater Agreement score of $0.4968$, as reported in the referenced paper~\cite{gezici2021evaluation}, indicate the subjective nature, thereby the difficulty of the stance classification task. The Fleiss-Kappa score~\footnote{Fleiss Kappa is a statistical measure which is an extended version of Cohen's Kappa (only for two raters) for assessing the reliability of agreement between a fixed number of raters when assigning categorical ratings to a number of items. This measure takes into account the agreement due to chance as well.} measures the agreement among multiple annotators for categorical ratings, while the Inter-Rater Agreement score quantifies the level of consensus among annotators. The relatively low scores in both metrics reflect the inherent subjectivity and challenges in accurately classifying stances.

\subsection{Stance Classification}
To examine the effect of multi-perspectivist approach on classification model performance, we investigate two different paradigms of including annotations to the model training phase, as~\emph{Baseline} and~\emph{Multi-perspective} models.

\paragraph{\textbf{Baseline Model}}
In the baseline model, we use the traditional label aggregation schema of~\emph{majority voting}, thus there is only one majority label per document. Based on this, in the baseline dataset, each document $d_i$ is composed of query, document content, and majority label defined as $d_i = \{q_i, c_i, m_i\}$.
\paragraph{\textbf{Multi-perspective Model}}
Unlike, for the multi-perspective model, we create the dataset through data augmentation by introducing more than one single label for each document. In fact, the annotation set for document $d_i$ is defined as $A(d_i)=\{a_1, a_2, a_3\}$, where each annotation potentially may vary depending on annotator's perspective.  
Thus, the multi-perspective dataset consists of $d_i$, where $d_i$ is added to the dataset three times with the corresponding annotations as ${d_i}^1 = \{q_i, c_i, {a_1}\}$, ${d_i}^2 = \{q_i, c_i, {a_2}\}$, and ${d_i}^3 = \{q_i, c_i, {a_3}\}$.

\paragraph{\textbf{Chunking}}
Since the transformer-based models have maximum length limitation, which is 512 for the base models of BERT and RoBERTa, the content is truncated to 512, if there is no chunking. In the case of chunking  which is a widely used technique for handling long inputs, the input is splitted into segments (chunks) using the library of chunkipy\footnote{https://pypi.org/project/chunkipy/}. The chunkipy library ensures to provide complete and syntactically correct sentences through using the stanza library~\cite{qi2020stanza} with a flexible overlapping to preserve the context along the chunks. Note that the chunks might have different lengths.

\begin{figure}[t]
    \centering
    \begin{subfigure}[b]{0.49\textwidth}
        \centering
        \includegraphics[width=\textwidth]{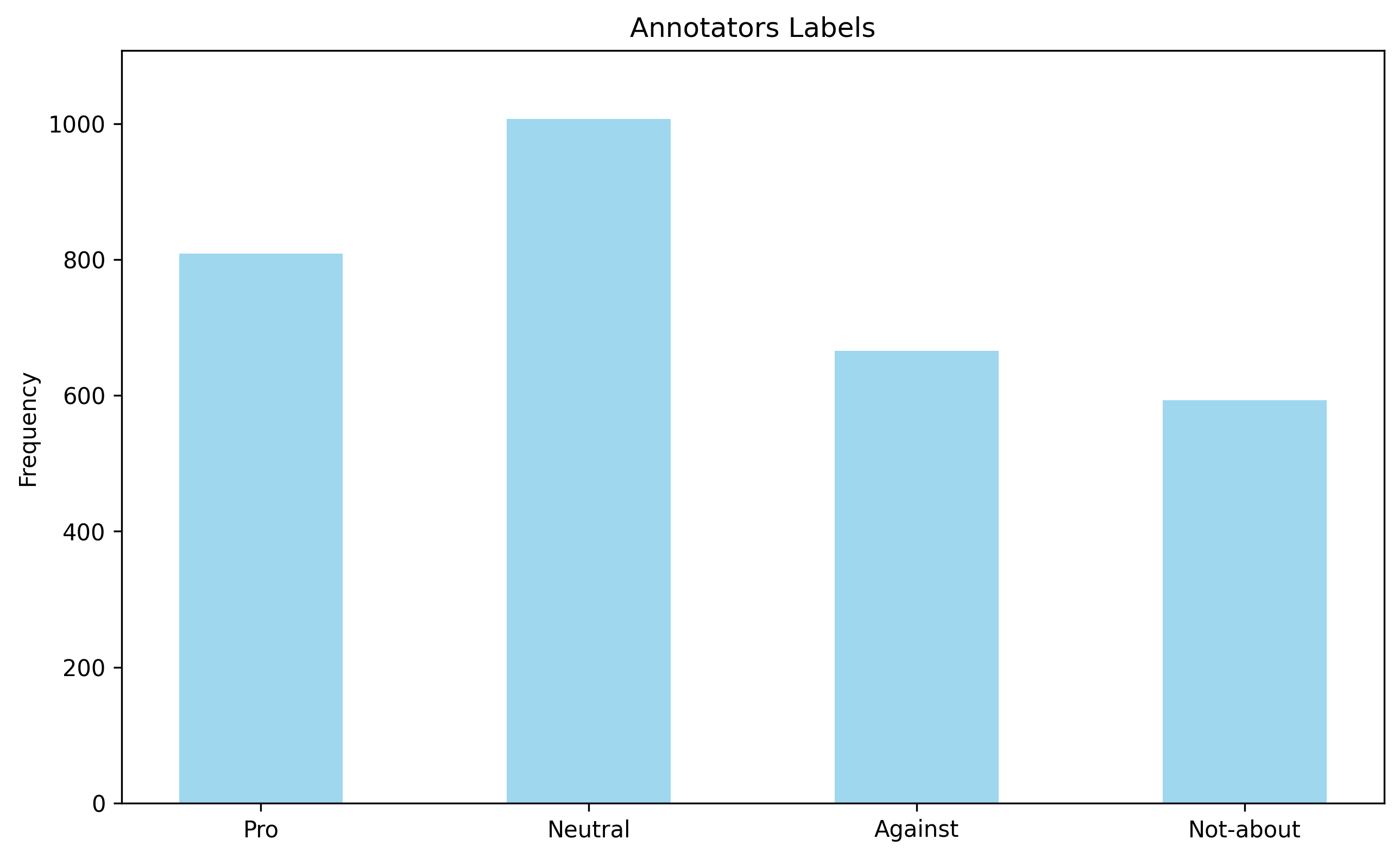}
        \caption{Original label distribution}
        \label{fig:fig1}
    \end{subfigure}
    \hfill
    \begin{subfigure}[b]{0.49\textwidth}
        \centering
        \includegraphics[width=\textwidth]{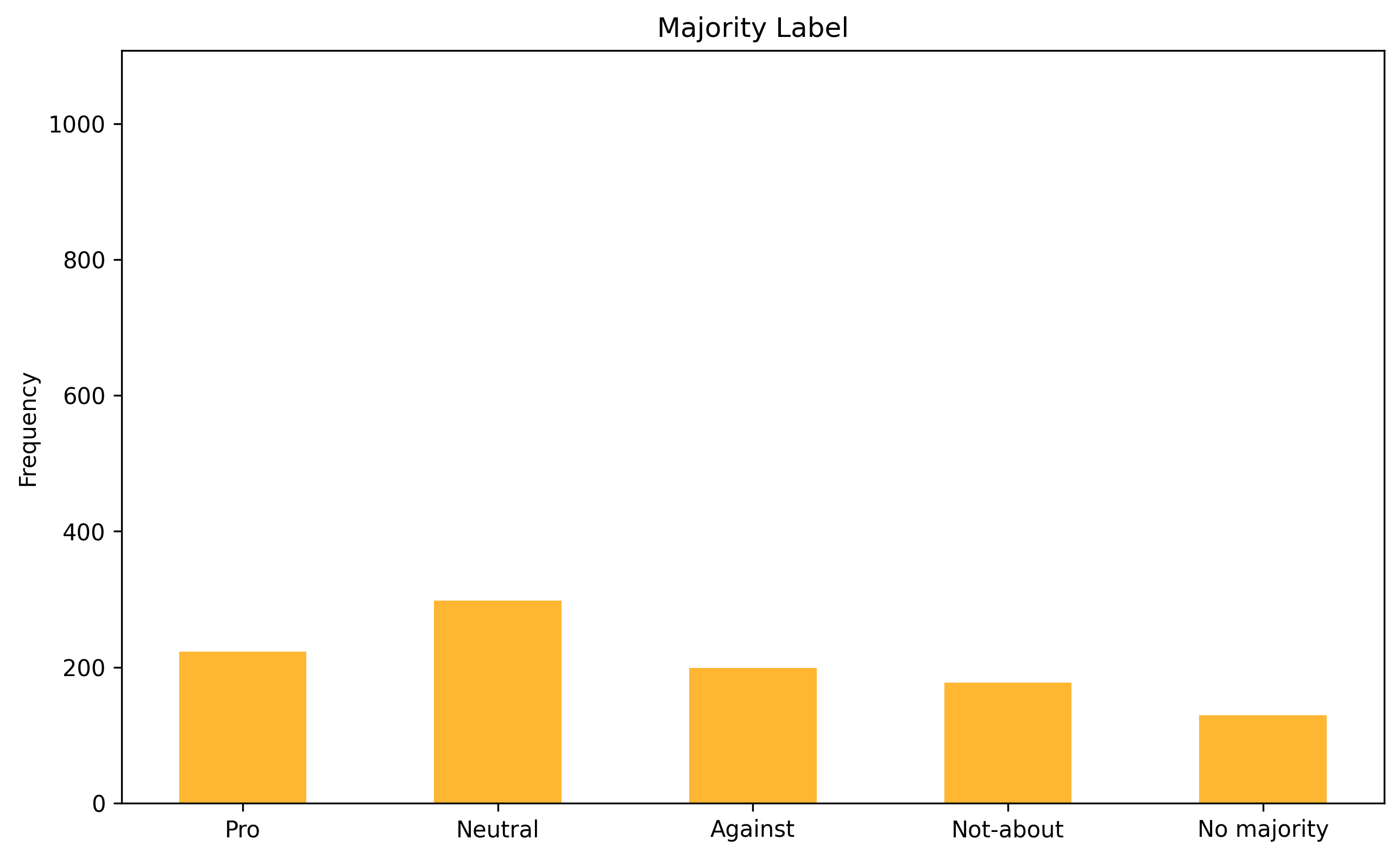}
        \caption{Majority label distribution}
        \label{fig:fig2}
    \end{subfigure}
    \label{fig:both}
\end{figure}

\section{Experimental Setup}
This section provides the description of our experimental setup based on the proposed methodology \footnote{Our code is publicly available at~\url{https://github.com/praveensonu/Multi-perspective-Stance-Detection}}. 

\subsection{Dataset Preprocessing}
The dataset consists of documents with varying lengths and some of them are extremely long\footnote{Probably due to scraping errors, 54 documents in the corpus contained 4 million tokens and when we manually checked their URLs, they were currently unavailable.}. Thus, we checked the document lengths and set the maximum length as 8,000 tokens since the majority of the documents (75\%) fit this condition, and further discarded the 54 extremely long documents.
Before applying the multi-perspective approach, 53 duplicate documents were eliminated as well.
Regarding the labels, we removed all the documents annotated with~\emph{link not-working}, i.e. 54 documents, as they do not provide any information for the stance classification task.
After removing these documents the dataset size became 1062. The original label distribution of the dataset is depicted in Figure~\ref{fig:fig1} and the label distribution after applying the majority voting is shown in Figure~\ref{fig:fig2}.
For the fine-tuning, the dataset has been split into train (75\%), validation (15\%), and test sets (15\%). Different datasets were used for each model: 
\begin{itemize}
    \item Baseline model: Dataset contains instances and each instance is composed of a query, document content, and majority label.
    \item Multi-Perspective model: Document duplication approach was used to prevent overfitting due to the small dataset size. The dataset contains instances and each instance repeated three times consisting of a query, document content and the corresponding annotation label.
\end{itemize}
Note that for both the baseline and multi-perspective models, model evaluation is fulfilled using the majority labels.
The multi-perspective approach was applied only to the training and validation sets, while the same test set (after majority voting applied) was used for the model evaluation of the both multi-perspective and baseline models.

Apart from these, regarding the longer documents, we also experimented with LongFormer~\cite{beltagy2020longformer} that can handle longer inputs. Yet, due to its poor performance, we decided to discard it and proceed with chunking for both baseline and multi-perspective models. 

\begin{table}[!t]
\centering
\renewcommand{\arraystretch}{1.0}
\begin{tabular}{clcccccc}
\toprule
\textbf{Approach} & \textbf{Model} & \textbf{Chunking} & \textbf{Acc.} & \textbf{Prec.} & \textbf{Rec.}& \textbf{F1} & \textbf{Avg. Conf.}\\ 
\midrule
\hline
\multirow{4}{*}{Baseline} &  \multirow{2}{*}{BERT-base} & no &  28.66 & 27.59 & 22.42 & 17.17 & 0.33\\ 
& & yes & 33.12 & 30.70 & 28.17 & 26.67 & 0.44\\
& \multirow{2}{*}{RoBERTa-base} & no & 36.30 & 34.99 & 31.82 & 27.07 & 0.39\\
& & yes & 45.85 & 39.47 & 43.13 & \textbf{40.48} & \textbf{0.52}\\
\hline
\multirow{4}{*}{Multi-Perspective} & \multirow{2}{*}{BERT-base} & no & 32.48 & 31.12 & 28.22 & \underline{24.81} & \underline{0.51}\\ 
& & yes & 47.48 & 53.90 & 49.86 & \underline{\textbf{50.21}} & \underline{0.52}\\ 

& \multirow{2}{*}{RoBERTa-base} & no & 47.77 & 44.27& 43.63& \underline{41.43} & \underline{\textbf{0.55}}\\ 
& & yes & 47.48 & 52.68 & 50.14 & \underline{47.45} & \underline{0.54}\\\hline
 
\bottomrule
\end{tabular}
\caption{Overall model evaluation results for the baseline and multi-perspective models}

\label{tab:results}
\end{table}

\subsection{Results}

The model evaluation results on the test set after the fine-tuning for the encoder-based models of BERT-base and RoBERTa-base~\cite{devlin-etal-2019-bert, liu2019roberta} are shown in Table~\ref{tab:results}. The fine-tuning has been fulfilled with a batch size of 32 for four epochs using the default hyperparameters on a 32GB Tesla V100 GPU.
For the evaluation, we reported different metrics as accuracy, precision, recall and F1 score, yet we have fulfilled the comparative evaluation based on the F1-score since the dataset is slightly imbalanced towards~\emph{neutral}. Apart from these evaluation metrics, we also reported the model confidence scores which were computed with a weighted average for models with chunking by taking into account the chunk length as well. Based on the results, multi-perspective models outperform the baseline models and chunking works better than no chunking as expected, since with chunking we provide more information to the model. These findings are applicable to the both BERT-base and RoBERTa-base.
The best-performing baseline model is RoBERTa-base with chunking, whereas the best multi-perspective model is BERT-base with chunking, although the RoBERTA-base with chunking has a comparable performance ($50.21$ vs. $47.45$) as well. This is probably due to the small dataset size or the effect of chunking.
Despite the slightly better performance of BERT over RoBERTa with chunking in the context of the multi-perspective approach, RoBERTa is more confident in prediction and this applies to all the results in Table~\ref{tab:results}. Also, regardless of the model, the models are more confident with chunking in both the baseline and multi-perspective settings.

\section{Conclusion \& Future Work} 
In this work, we present a methodology to incorporate multi-perspective models into the stance detection task.
To enhance the responsibility and ethical standards of NLP systems, we advocate for the implementation of perspectivist approaches and the use of disaggregated datasets. We believe these strategies are essential for the advancement of the field.
We used two transformer-based models as BERT-base and RoBERTa-base, i.e. one of the first LLMs, for our experimental setup and further evaluated them with and without chunking due to the sequence length limitation.
Our initial studies demonstrate that multi-perspective models outperform the baseline models with and without chunking regardless of the model itself.
Moreover, we observed that chunking improves model performance as well as the model confidence for both the baseline and multi-perspective settings.
These results highlight that developing more inclusive perspective-aware AI models improves the classification model performance in subjective NLP tasks.
In the scope of this study, we did not experiment with the larger versions of BERT and RoBERTa which we left as future work. Another potential future direction is to fulfill hyperparameter tuning and also for the data augmentation phase, to add a document to the dataset three times only if the given document has three different annotations, i.e. exploring different data augmentation techniques.
We plan to have a more detailed analysis, e.g. query-wise, related to the annotator disagreement level on model performance as well as confidence scores. Finally, a similar methodology could be applied on~\emph{ideology detection} which is the second main task in the original paper of the dataset.

\section*{Acknowledgements} 
This work has been supported by the European Union under ERC-2018-ADG
GA 834756 (XAI), by HumanE-AI-Net GA 952026, and by the Partnership Extended PE00000013 - “FAIR - Future Artificial Intelligence Research” - Spoke 1 “Human-centered AI”.This work has been partially supported by the EU H2020 TAILOR project, GA n. 952215.

\bibliography{resource.bib}


\end{document}